\documentstyle[11pt,epsfig]{article}

\newcommand{\bd}{\begin{document}}
\newcommand{\ed}{\end{document}}
\newcommand{\bc}{\begin{center}}
\newcommand{\ec}{\end{center}}
\newcommand{\vs}{\vspace}
\newcommand{\hs}{\hspace}
\newcommand{\bq}{\begin{quote}}
\newcommand{\eq}{\end{quote}}
\newcommand{\lb}{\linebreak}
\newcommand{\mb}{\makebox}
\newcommand{\mc}{\multicolumn}
\newcommand{\bit}{\begin{itemize}}
\newcommand{\eit}{\end{itemize}}
\newcommand{\ben}{\begin{enumerate}}
\newcommand{\een}{\end{enumerate}}
\newcommand{\ol}{\overline}
\newcommand{\lt}{\left}
\newcommand{\rt}{\right}
\newcommand{\hf}{\hspace*{\fill}}
\newcommand{\vf}{\vspace*{\fill}}
\newcommand{\beq}{\begin{equation}}
\newcommand{\eeq}{\end{equation}}
\newcommand{\ba}{\begin{array}}
\newcommand{\ea}{\end{array}}
\newcommand{\beqa}{\begin{eqnarray}}
\newcommand{\eeqa}{\end{eqnarray}}
\newcommand{\beqas}{\begin{eqnarray*}}
\newcommand{\eeqas}{\end{eqnarray*}}
\newcommand{\bfg}{\begin{figure}}
\newcommand{\efg}{\end{figure}}
\newcommand{\pad}{\partial}

\newcommand{\bm}[1]{\mb{\boldmath ${#1}$}}
\newcommand{\fb}[1]{\lt[{#1}\rt]}
\newcommand{\ot}{\otimes}
\newcommand{\nn}{\nonumber}
\newcommand{\<}{\langle}
\renewcommand{\>}{\rangle}
\newcommand{\C}{{\cal C}}
\newcommand{\M}{{\cal M}}
\newcommand{\F}{{\cal F}}

\newcommand{\Q}{{\cal Q}}
\newcommand{\N}{{\cal N}}
\newcommand{\A}{{\cal A}}
\renewcommand{\P}{{\cal P}}
\newcommand{\es}{\emptyset}
\newcommand{\ci}{\subseteq}
\newcommand{\cs}{\supseteq}
\renewcommand{\u}{\cup}
\renewcommand{\i}{\cap}
\newcommand{\bu}{\bigcup}
\newcommand{\bi}{\bigcap}
\newcommand{\la}{\leftarrow}
\newcommand{\ra}{\rightarrow}
\newcommand{\Ra}{\Rightarrow}
\newcommand{\Lra}{\Leftrightarrow}
\newcommand{\lgra}{\longrightarrow}
\newcommand{\Lgra}{\Longrightarrow}
\newcommand{\lglra}{\longleftrightarrow}
\newcommand{\Lglra}{\Longleftrightarrow}
\renewcommand{\S}{{\cal S}}
\renewcommand{\a}{\alpha}
\renewcommand{\b}{\beta}
\newcommand{\g}{\gamma}
\newcommand{\G}{\Gamma}
\renewcommand{\d}{\delta}
\newcommand{\th}{\theta}
\newcommand{\ph}{\phi}
\newcommand{\e}{\varepsilon}
\newcommand{\eps}{\epsilon}
\newcommand{\h}{\eta}
\renewcommand{\l}{\lambda}
\newcommand{\m}{\mu}
\newcommand{\n}{\nu}
\newcommand{\p}{\pi}
\newcommand{\s}{\sigma}
\newcommand{\Si}{\Sigma}
\newcommand{\ta}{\tau}
\newcommand{\Ph}{\Phi}
\renewcommand{\c}{\chi}
\newcommand{\om}{\omega}
\newcommand{\Om}{\Omega}
\newcommand{\tri}{\triangle}
\newcommand{\rec}[1]{\frac{1}{#1}}
\newcommand{\f}{\frac}
\newcommand{\sm}[2]{\sum_{#1}^{#2}}
\newcommand{\ld}{\ldots}
\newcommand{\ov}{\overline}
\newcommand{\un}{\underline}
\newcommand{\iy}{\infty}
\newcommand{\qed}{\hf\rule{3mm}{3mm}}
\newcommand{\wt}{\widetilde}
\newcommand{\ds}{\displaystyle}
\newcommand{\bdm}{\begin{displaymath}}
\newcommand{\edm}{\end{displaymath}}
\newcommand{\et}[2]{#1_{1},#1_{2},\ld  ,#1_{#2}}
\newcommand{\alter}[2]{\lt\{ \ba {ll}#1 \\ #2 \ea \rt.}
\newcommand{\altt}[3]{\lt\{ \ba {lll}#1 \\ #2 \\ #3 \ea \rt.}
\newcommand{\alt}[4]{\lt\{ \ba{ll}#1 & \mb{if \,\,}#2 \\ #3 & \mb{if
               \,\,}#4 \ea \rt.}
\newcommand{\altn}[4]{\lt\{ \ba{rl}#1 & \mb{if \,\,}#2 \\ #3 & \mb{if
               \,\,}#4 \ea \rt.}
\newcommand{\alto}[6]{ \lt\{ \ba{ll}#1 & \mb{if \,\,}#2 \\ #3 & \mb{if
               \,\,} #4 \\ #5 & \mb{if \,\,}#6 \ea \rt.}
\newcommand{\altero}[5]{\mb{$\lt\{ \ba {ll}#1 & \mb{if \,\,}#2 \\ #3 &
               \mb{if \,\,} #4 \\ #5 & \mb{otherwise} \ea \rt.$}}
\newcounter{algc}
\newcounter{romc}
\newcounter{alphc}
\newcommand{\bl}{\begin{list}{{\it Step}~\arabic{algc}~:} {\usecounter{algc}
                        \setlength{\topsep}{0pt} \setlength{\itemsep}{0pt}}}
\newcommand{\el}{\end{list}}
\newcommand{\blr}{\begin{list}{~\roman{romc}~)} {\usecounter{romc}
                        \setlength{\topsep}{0pt} \setlength{\itemsep}{0pt}}}
\newcommand{\elr}{\end{list}}
\newcommand{\bla}{\begin{list}{~\alph{alphc}~)} {\usecounter{alphc}
                        \setlength{\topsep}{0pt} \setlength{\itemsep}{0pt}}}
\newcommand{\ela}{\end{list}}

\bd
\setlength{\parskip}{0.5pc}
\setlength{\parindent}{0.0cm}
\addtolength{\oddsidemargin}{-2cm}
\addtolength{\topmargin}{-1.5cm}

\rm
\pagenumbering{arabic}

\title{{\huge {\bf  Designing fuzzy rule based  classifier
using   self-organizing feature map for
  analysis of  multispectral satellite images\thanks{International Journal of Remote Sensing, Volume 26, No 10, Pages 2219-2240, May 2005.}}}}

\author{Nikhil R. Pal\thanks{Electronics and Communication Sciences Unit,
Indian Statistical Institute,
203 B. T. Road, Calcutta 700 108.}, Arijit Laha\thanks{Institute for Development and Research in Banking Technology,
Castle Hills, Masab Tank, Hyderabad 500 057} and J. Das\thanks{Electronics and Communication Sciences Unit, Indian Statistical Institute,
203 B. T. Road, Calcutta 700 108.}
}

\date{}
\maketitle
{\normalsize{\bf Abstract}}

{\small
We propose a novel scheme for designing fuzzy rule based classifier. An
SOFM based method is used for generating a set of prototypes which is used to
generate a set of fuzzy rules. Each rule represents a region in the feature
space that we call the context of the rule. The rules are tuned with respect
to their context. We justified that the reasoning scheme may be different in
different context leading to context sensitive inferencing. To realize
context sensitive inferencing we used a softmin operator with a tunable
parameter. The proposed scheme
is tested on several multispectral satellite image data sets and the performance is
found to be much better than the results reported in the literature.}

\section{Introduction}

A classifier \cite{dh1}
can be defined as any function ${\cal D}:\Re^{p} \rightarrow N_{c}$,
where $N_{c}=\{{\bf e}_{i}
\mid i=1,...,c,{\bf e}_{i} \in \Re^c\}$ is the set of label vectors and $c$ is the number
of classes.  Given any feature vector  ${\bf x} \in  \Re^{p} $, it produces a label
vector in $\Re^{c}$.
If ${\cal D}$
is a crisp classifier, ${\bf e}_{i}$s are basis vectors with components $e_{ij} = 0 \forall
i \neq j$ and $e_{ii} = 1$. If ${\cal D}$ is a fuzzy classifier then $e_{ij} \geq 0$ and
$ \sum_{j=1}^c e_{ij} = 1 $. Designing a classifier involves finding a good
${\cal D}$.

Although Bayes' classifier is optimal \cite{dh1}
 it  requires statistical knowledge of the sample set
in terms of  prior probabilities and  class conditional densities,
which are almost never available in practical cases. Usually no
knowledge of the underlying distribution is available except
whatever can be inferred from   samples. In such a case the
classifier has to rely  on a set of labeled samples $X=\{({\bf
x}_{i},{\bf l}_{i}\mid i=1,...,n, {\bf x}_{i} \in \Re^{p}, {\bf
l}_{i} \in N_{c})\}$, where ${\bf l}_{i}$ is the label vector associated with
 the
training vector ${\bf x}_{i}$.  The training samples are assumed to be a true
representative of the data to be classified. Nonparametric
classification schemes do not use any parametric model \cite{gong1,gong2}
but utilize the training samples in a suitable training procedure
to model the distribution of the input data and use the model to
infer about the class membership of unknown data points.

For example, a crisp $k$-NN classifier
uses the whole set of training samples to infer the
class membership of an unknown input vector ${\bf x} \in \Re^{p}$ using the following
rule:\hf\lb
{\em $\diamond$ Find the set of $k$ samples $\{({\bf x}_{i},{\bf l}_{i})\} \in X$
closest to the sample ${\bf x}$ \hf\lb
$\diamond$ Assign the sample ${\bf x}$ to the class from which majority of $k$
closest neighbors has come.\hf\lb}

On the other hand, a prototype based classifier approximates the distribution
of the training data through a set of prototypes
$V=\{{\bf v}_{i},{\bf l}_{i} \mid i=1,...,\hat{c}\geq c,{\bf v}_{i} \in \Re^{p},
{\bf l}_{i} \in N_{c}\}$, and classifies a vector ${\bf x} \in \Re^{p}$
as follows.\hf\lb
\centerline{Decide ${\bf x} \in$ class $i \Leftrightarrow {\cal D}_{V,\d}({\bf x}) = {\bf l}_{i} \Leftrightarrow
\d({\bf x},{\bf v}_{i}) \leq \d({\bf x},{\bf v}_{j}) \forall i \neq j .$}\hf\lb
where $\d$ is usually the Euclidean distance function.

In recent times fuzzy rule based classifiers \cite{ bez1, bez2,
ishi1, ludmila} have attracted attention of many researchers due
to their several attractive features compared to more traditional
distance based classifiers. At the conceptual level their working
is closer in spirit to human reasoning. At the practical level
they can be used very easily to handle several problematic
situations. For example, detection of outliers can be done using
small firing strengths of the rules, highly overlapped regions can
be detected by high firing strength for rules from more than one
class. Most interestingly, the problem of high variation in the
variances of different features, which often degrades the
performance of a distance based classifier substantially, can be
handled naturally by fuzzy rules due to the atomic nature of the
antecedent clauses. It may be noted that the use of Mahalanobis
distance can handle, to some extent, the problem  associated with
variation in variances of different features. However, if the
training data from a class are divided into several clusters with
different distributions, then Mahalanobis distance may not be very
effective, but fuzzy rules will be able to model such class
structures in a natural manner.

A fuzzy rule based classifier consists of a set of fuzzy rules of the form:\hf\lb
$R_i:$ {\bf If $x_1$ is $A_{i1}$ AND $x_2$ is $A_{i2}$ AND $\cdots$ AND
$x_p$ is $A_{ip}$ then class is $j$}.\hf\lb
Here $A_{ik}$ is a fuzzy set used in the i-th rule  and
defined on the domain of $x_k$, i.e., on the universe of the k-th feature.

When a sample data point ${\bf x} \in \Re^{p}$ is presented to the system for
classification, the
fuzzy rules fire to produce outputs. The magnitude of the outputs ( also known
as {\em firing strengths}) are used for deciding the class membership of the
sample data ${\bf x}$.

For designing a fuzzy rule based classifier there are three issues that
need to be addressed:

$I_1$ : How many rules are needed?\hf\lb
$I_2$ : How to generate the rules? \hf\lb
$I_3$ : How to use the rules to decide a class? \hf\lb

The simplest way of tackling the above issues is to take the help of a domain
expert and create the fuzzy rules to represent his/her domain knowledge. But in
a typical pattern classification problem, such domain knowledge is usually not
available. So a scheme is needed for designing fuzzy rule based classifier
based on the training samples.

In this paper we describe a comprehensive scheme for designing fuzzy
rule based classifiers. The scheme takes care of all the three issues mentioned
above. This is a multi-stage scheme. In the first stage a set of
labeled prototypes representing the distribution of the training data is
generated using a Self-organizing Feature Map (SOFM) \cite{k1,laha,laha2}.
The algorithm employs a combination
of unsupervised and supervised clustering of the training data to generate an
adequate number of prototypes representing the overall as well as class-specific
distribution of the training data. Then each of these prototypes is converted
to a fuzzy rule of the form described above. Thus, each of the fuzzy rules
represents a region, may be overlapped, in the feature space. We call this
region the {\bf context} of the rule. Note that, throughout this paper the
word ``context" is used to mean a region in the feature space, not pixels
in the neighborhood of another pixel in an image.

Next we develop a tuning algorithm for the fuzzy rules that fine-tunes
the peaks as well as the spreads of the fuzzy sets associated with the rules.
We call this the {\em context tuning} stage.
However, the exact implementation of the tuning algorithm is dependent
on the conjunction operator used to represent the AND connective in the
antecedent part of the rules. Here we derive tuning algorithms for
two different conjunction operators, namely the {\bf product} and the
{\bf softmin}. The tuned rules are used to classify the unknown samples
based on the firing strengths of the rules. A test sample is classified to
the class of the rule generating the highest firing strength.

The softmin operator, based on the value of a parameter can approximate
a whole family of operators including {\em min, average} and {\em max}.
This raises the possibility of using a set of rules where each rule
is free to use a different conjunction operator depending on the context
it operates on. This is known as {\bf context-sensitive reasoning}. Here,
depending on the context the reasoning scheme may change. To
realize this possibility we also develop an algorithm for tuning the
parameters of the softmin operators on a per-rule basis.
The schemes using the same conjunction operator (reasoning scheme)
for all rules will be called {\bf context-free reasoning} in this paper.

Classification of multispectral satellite images is a very
important field of application of pattern recognition
techniques. Currently huge amount of
information about the earth is routinely being generated by the satellite-based sensors.
This information is often available in the form of multispectral images produced by a set
of sensors operating in different spectral regions. A sensor operating in certain
spectral region might be more sensitive to certain classes of objects than the others.
Hence, to develop a good analysis system it is necessary to use
 data available from as many
sensors as possible. This requirement led to development of numerous
techniques for data fusion and classification.
To name a few, statistical methods \cite{solberg, paola}, Dempster-Shafer
theory \cite{lee}, neural networks \cite{paola, ben, atkin, pinz} etc.

Numerous fuzzy classification techniques have been developed by
many researchers to solve problems in various fields. A
comprehensive account of such works can be found in \cite{bez1,
bez2, ludmila}. Many references are available on the use of
different fuzzy methodologies for land cover classification from
multispectral satellite images. For example, \cite{foody, cannon}
uses fuzzy $c$-means algorithm \cite{bez1}, Kumar et al.
\cite{kumar2} applied fuzzy integral method. Fuzzy rule base has also
been used for classification by many researchers \cite{bez2,
ishi1} for diverse fields of application. Fuzzy rules are
attractive because they are interpretable and can provide an
analyst a deeper insight into the problem. Not many attempts have been
made to use fuzzy rule based
systems for land cover analysis. In a
recent paper B\'{a}rdossy and Samaniego \cite{bard} have proposed
a scheme for developing a fuzzy rulebased classifier for analysis
of multispectral images. They employed simulated annealing for
optimizing the performance of a randomly selected initial set of
rules. In the present paper we propose a prototype-based fuzzy
rule generation approach \cite{bez2}, where the number of
prototypes (hence the number of rules)
depends on the complexity of the training data.

Though our scheme is applicable to any pattern recognition task
using object data, we have chosen to test our scheme on satellite
images because context sensitive inferencing could be very
effective for them. Analysis of satellite images has many
important applications such as prediction of storm and rainfall,
assessment of natural resources, estimation of crop yields,
assessment of natural disasters, and land cover classification. In
this paper we focus on land cover classification from
multi-spectral satellite images. We consider a set of independent
detectors of a sensor, operating in different spectral bands and
producing homogeneous data (i.e., same type of information, namely
pixel values). The weakly coupled fused data \cite{clark} consist
of one data vector for each pixel, data from each detector
contributing to one dimension of the data vectors. We use the
weakly coupled data along with the ground truth to train a
classifier. We  use images from two types of sensors, a 7-channel
image and a 4-channel  image produced by a Landsat  Multispectral
Scanner \cite{satim} and a Thematic Mapper respectively
\cite{ludmila}.

Details of the scheme and the experimental results are described in the following
sections. Section 2 covers the method of generating the prototypes,
the method for converting the
prototypes into fuzzy rules and the context tuning algorithms. Section 3 contains
the description of context sensitive reasoning methods. The experimental results
and the discussions are presented in section 4. Section 5 concludes the paper.

\section{Designing of the Fuzzy Rule based Classifiers}
The proposed scheme has several stages. We use Kohonen's Self-Organizing Feature
Map (SOFM)\cite{k1} to obtain a set of prototypes. For the sake of completeness
we provide a brief description of SOFM.

\subsection{Kohonen's SOFM algorithm}
The self-organizing feature map is an algorithmic
transformation denoted here by $A^{D}_{ SOFM} : \Re^{p} \ra V(\Re^{q})$
that is often advocated for visualization of metric-topological
relationships and distributional density properties of feature vectors
(signals) $X = \{{\bf x}_{1},...,{\bf x}_{N}\}$ in $\Re^{p}$. SOFM is implemented through a
two-layered neural  architecture that is believed to be similar in some ways
to the biological neural network. The input layer is a fan-out layer and the
output layer is a competitive layer. Each node in the input layer
is connected to all nodes in the competitive layer.

The visual display produced by $A^{D}_{SOFM}$ presumably helps one
form hypotheses about topological structure in $X$. In principle
$X$ can be transformed onto a display lattice in $\Re^{q}$ for any
$q$; in practice, visual display can be made only for $q \leq 3$
and are usually made on a linear or planar configuration arranged
as a rectangular or hexagonal lattice. Here we explain
architecture and training procedure of SOFM using ($m \times n$)
display or output nodes.

Input vectors ${\bf x} \in \Re^{p}$ are distributed by a fan-out layer to each
of the ($m \times n$) output nodes in the competitive layer. Each node
in this layer has a weight vector (prototype) ${\bf w}_{ij}$ attached to it.
We let $ O_{p} = \{{\bf w}_{ij}\} \subset \Re^{p}$
denote the set of ${\em m}\times{\em n}$ weight vectors. $O_{p}$ is
(logically) connected to a display grid $O_{2} \subset V(\Re^{2})$.
($i,j$) in the index set $\{1,2,\ldots, m\}\times\{1,2,\ldots,n\}$
is the logical address of the cell. There is a one-to-one
correspondence between the $m \times n$ {\em p}-vectors {${\bf w}_{ij}$}
and the $m \times n$ cells ($\{i,j\}$),i.e., $O_{p} \lglra O_{2}$.

SOFM usually begins with a    random initialization of the weight vectors
{${\bf w}_{ij}$}. For notational clarity we
suppress the double subscripts. Now let ${\bf x} \in \Re^{p}$
enter the network and let $t$ denote the current iteration number. Find
${\bf w}_{r,t-1}$, that best matches ${\bf x}$ in the sense of minimum
Euclidean distance in $\Re^{p}$. This vector has a (logical) ``image" which is
the cell in $O_{2}$ with subscript $r$. Next a topological (spatial)
neighborhood $N_{r}(t)$ centered at $r$ is defined in $O_{2}$, and its
display cell neighbors are located. For example, $3 \times 3$ window,
$N(r)$, centered at $r$ corresponds to   nine prototypes in
$\Re^{p}$. Finally, ${\bf w}_{r,t-1}$ and the other weight vectors associated with
cells in the spatial neighborhood $N_{t}(r)$ are updated using the rule
\beq {\bf w}_{i,t} = {\bf w}_{i,t-1} + h_{ri}(t)({\bf x} - {\bf w}_{i,t-1}) .\eeq
Here $r$ is the index of the ``winner" prototype
\beq r = \underbrace{arg\ min}_{i} \{\|{\bf x} - {\bf w}_{i,t-1}\|\} \eeq
and $\|*\|$ is the Euclidean norm on $\Re^{p}$.
The function $h_{ri}(t)$ which expresses the strength of
interaction between cells $r$ and $i$ in $O_{2}$, usually decreases
with $t$, and for a fixed $t$ it decreases as the distance (in $O_{2}$) from cell
$r$ to cell $i$ increases. $h_{ri}(t)$ is usually expressed as the product
of a learning parameter $\a_{t}$ and a lateral feedback function $g_{t}(dist(r,i))$.
 A common choice for $g_{t}$ is
$g_{t}(dist(r,i))=\exp^{-dist^{2}(r,i)/\s^{2}_{t}}$. $\a_{t}$ and $\s_{t}$ both
decrease with time $t$. The topological neighborhood $N_{t}(r)$ also decreases
with time. This scheme when repeated long enough, usually preserves spatial order
in the sense that weight vectors which are metrically close in $\Re^{p}$ generally
have, at termination of the learning procedure, visually close images on the display
lattice.

\subsection{Generation of Prototypes}
We use a 1-D SOFM, but the algorithm can be extended to
2-D SOFM also. First we train a one-dimensional SOFM using the training data,
of course, without using the class
information of the input data.  We start the SOFM with c nodes where
$c$ is the number of classes. We do so because
the smallest number of rules that may be required is equal to the number of
classes. At the end of the training
the weight vector distribution of the SOFM   reflects the distribution of the
input data. These unlabeled prototypes are then labeled
using class information. For each of $N$ input feature vectors we
identify the prototype closest to it, i.e., the winner node. Since no class
information is used during
the training,   some prototypes may become the winner
for data from more than one class. For each prototype ${\bf v}_{i}$
we compute a score $D_{ij}$, which is the number of data points from class $j$ to
which ${\bf v}_{i}$ is the closest prototype. If $D_{ij}$ is 0 for
all $j$ for a prototype
${\bf v}_{i}$   then we reject it.  For the remaining prototypes the
class label $C_{i}$ of the prototype ${\bf v}_{i}$ is determined as
\beq C_{i} = \underbrace{arg \max}_{j} D_{ij} .\eeq
The scheme   assigns a label to each of the $c$ prototypes, but such a set
of prototypes may not classify the data satisfactorily. For example, from
(3) it is clear that $\sum_{j \neq i}D_{ij}$ data points will be wrongly
classified by the prototype ${\bf v}_{i}$. Hence we need further refinement of
the initial set of prototypes
$V_{0} = \{ {\bf v}_{10},{\bf v}_{20}, ..., {\bf v}_{c0}\} \subset \Re^{p}$.

We use the prototype refinement scheme described in \cite{laha,laha2}. The basic idea
behind this refinement algorithm is that a useful prototype, ${\bf v}_i$
should satisfy two criteria:\hf\lb
(i) It should represent adequate number of points, i.e., $W_i=\sum_{j=1}^cD_{ij}$
should be high.\hf\lb
(ii) Only one of the classes, say class $k$, should be strongly represented by
${\bf v}_i$, i.e., $D_{ik}$ should be high and $D_{ij}, \forall j \neq k$ should
be low.

If condition (i) is not satisfied ${\bf v}_i$ is deleted and depending on the
values of $D_{ij}$ prototypes are split and merged. Finally, the set of prototypes
are again refined by SOFM algorithm with winner-only update strategy. After a
few iterations this algorithm produces a set of adequate number of prototypes
that represents the training data much better than the initial one. For details the
readers are referred to \cite{laha,laha2}.

Now we use these prototypes to generate fuzzy rules that we describe next.

\subsection{Designing fuzzy rulebased classifiers}
A prototype (representing a cluster of points) ${\bf v}_i$ for   class $k$
can be translated into a fuzzy rule of the form :

\bc
$R_{i}$: If ${\bf x}$ is CLOSE TO ${\bf v}_i$ then the class is $k$.
\ec

Where the fuzzy set ``CLOSE TO" can be represented by a multidimensional
membership function such as
$$\mu_{CLOSE TO}({\bf x}) = \exp^{-\frac{||{\bf x}-{\bf v}_i||^2} {{\sigma_i}^2} },$$
where $\sigma_i > 0$ is a constant. This is equivalent to using prototypes
with hyperspherical zones of influence centered at ${\bf v}_i$s. The 1-MSP
(most similar prototype) classifier uses such membership values and it
has been studied in \cite{laha2}. Such a classifier does not perform quite
well when different features have considerably different variances.

To overcome this shortcoming, ``${\bf x}$ is CLOSE TO ${\bf v}_i$"
can be written as a conjunction of $p$ atomic clauses :
\bc
$x_1$ is CLOSE TO $v_{1}$ AND $\cdots$ AND $x_p$ is CLOSE TO $v_{p}$.
\ec
Such that the i-th rule $R_{i}$ representing one of the c classes takes the form
\bc
$R_{i}$ : $x_1$ is CLOSE TO $v_{i1}$ AND $\cdots$ AND $x_p$ is
CLOSE TO $v_{ip}$ then class is $k$.
\ec
The fuzzy set   CLOSE TO $v_{ij}$  can be modeled by triangular, trapezoidal
or Gaussian membership function. In this investigation, we use the Gaussian
membership function,
$$\mu_{ij}  ( x_j; v_{ij},\sigma_{ij})=\exp{-{(x_j - v_{ij})}^2/{\sigma_{ij}}^2}.$$
Given a   data point ${\bf x}$ with unknown class, we first find
the firing strength of each rule. Let $\alpha_i({\bf x})$ denote
the firing strength of the $i^{th}$ rule on a data point ${\bf
x}$.
The firing strength can be computed using any
T-norm \cite{bez2}.  For example, using {\bf product}  the firing strength becomes
 $$\alpha_i({\bf x}) = \Pi_{j=1}^{j=p}\mu_{ij}(x_j; v_{ij}, \sigma_{ij}).$$
We then assign the point ${\bf x}$ to class $k$, if $\alpha_r =
\max_i(\alpha_i({\bf x}))$ and the $r^{th}$ rule represents class
$k$.

The performance of the classifier depends crucially on the adequacy of the number of
rules used and proper choice of the fuzzy sets used in the antecedent part of the rules.
In our case each fuzzy set is characterized by two parameters $v_{j}$ and $\sigma_{ij}$.
The $v_{ij}$s of the rules can be initialized with the components of the
final prototypes generated by our SOFM based algorithm, $V^0= V^{final}=\{{\bf v}_1^{final}, \cdots, {\bf v}_{\hat c}^{final} \}$
$=\{{\bf v}_1^{0}, \cdots, {\bf v}_{\hat c}^{0} \}$
 where $v_{ij}^0=v_{ij}^{final}$
or they can be generated using any clustering  algorithms like the
fuzzy c-means \cite{bez1} provided we know the required number of rules.
A distinct advantage of using  our SOFM based method is that
it automatically decides on the required number of rules.
 The initial estimates of the
$\sigma_{ij}$s are computed as follows.

For each prototype ${\bf v}_{i}^{0}$ in the set $V^{0}=\{{\bf v}_{i}^{0} \mid i=1,...,
\hat{c}, {\bf v}_{i}^{0} \in \Re^{p} \}$
let $X_{i}$ be the set of training data closest to ${\bf v}_{i}^{0}$. For each
${\bf v}_{i}^{0}$ the set
$$S_{i} = k_{w}\{ \sigma_{ij} \mid j=1,...,p, \sigma_{ij}
=(\sqrt(\sum_{{\bf x}_{k} \in
X_{i}}(x_{kj}-v_{ij}^0)^2))/|X_{i}|\}$$
is computed and associated with the prototype. $k_{w} > 0$ is a constant parameter
that controls the initial width of the membership function. Its value can
have a significant
impact on the classification performance for complicated data sets.

The initial rulebase $R^0$ thus obtained can be further fine tuned to achieve better
performance. But the exact tuning algorithm depends on the conjunction
operator (implementing AND operation for the antecedent part) used for computation
of the firing strengths. We mentioned earlier that he firing strength can be calculated using any
conjunction operator or T-norm \cite{bez2}. Use of different T-norms
result in different classifiers. The {\bf product} and the
{\bf minimum} are among most popular T-norms used as conjunction operators.
Using product the firing strength of r-th rule is computed as follows:
$$\alpha_r({\bf x}) = \Pi_{j=1}^{j=p}\mu_{rj}(x_j; v_{rj}, \sigma_{rj}).$$
and the same when computed using the minimum is
$$\alpha_r = \min_j\{\mu_{rj}(x_j; v_{rj}, \sigma_{rj})\}.$$
It is much easier to formulate a calculus based tuning algorithm if
product is used.

In the current study we design two different classifiers, one using product and
the other using the {\bf Soft-min} operator. We shall see that soft-min enables
us to realize a novel context sensitive inferencing scheme.

\subsubsection{Tuning of the Rule Base}
Let ${\bf x} \in X$
be from class $c$ and ${ R}_{c}$ be the rule from class $c$
giving the maximum firing strength
$\a_{c}$ for  ${\bf x}$.  X is the training set.
Also let ${R}_{\neg c}$ be the rule
 from
the incorrect classes having the highest firing strength
$ \a_{\neg c}$ for  ${\bf x}$.

We use the error function $E$,
\beq E = \sum_{{\bf x} \in X}(1-\a_{c}+\a_{\neg c})^{2}. \eeq

This error function has been used by Chiu \cite{chiu1}.
We minimize $E$ with respect to $v_{cj}$, $v_{\neg cj}$ and $\sigma_{cj}$,
$\sigma_{\neg cj}$ of the two rules $R_{c}$ and $R_{\neg c}$. This will
refine the rules with respect to their contexts.

Here the index $j$ corresponds to clause number in the corresponding rule,
i.e., for the first antecedent clause $j=1$, for the second clause
$j=2$ and so on.  Next we give an algorithmic
description of the rule refinement algorithm when product is used to compute
the firing strength.


{\em Rule refinement (context-tuning) algorithm:}\hf\lb

{\em Begin}\hf\lb

Choose learning parameters $\eta_{m}$ and $\eta_{s}$.\hf\lb
Choose a parameter reduction factor $ 0 < \e < 1 $.\hf\lb
Choose the maximum number of iteration $maxiter$.\hf\lb
Compute   the error   $E_{0}$ for the initial
rule base $R^{0}$.  \hf\lb
 Compute the misclassification $M_{0}$ Corresponding to initial
rule base $R^{0}$.\hf\lb

$t\leftarrow 1$\hf\lb While ($t \leq maxiter$) do\hf\lb
\hs*{1cm}For each  vector ${\bf x} \in X_{Tr}$ (The training
set)\hf\lb
\hs*{2 cm}Find the rules ${R}_{c}$ and ${ R}_{\neg c}$
 using  $\a_{c}$ and $\a_{\neg c}$.\hf\lb
\hs*{2 cm}Modify the parameters of rules  ${R}_{c}$ and ${R}_{\neg c}$
as follows .\hf\lb

\hs*{2 cm}For $k=1$ to $p$ do \hf\lb
\hs*{2.5 cm}(A) $v_{ck}^{new} =
v_{ck}^{old} - \eta_{m}\frac{\partial E}{\partial v_{ck}^{old}} =
v_{ck}^{old}+\eta_{m}(1-\a_{c}+\a_{\neg c})
\frac{\a_{c}}{\s_{ck}^{old^{2}}}(x_{k}-v_{ck}^{old})$\hf\lb
\hs*{2.5 cm}(B) $v_{\neg ck}^{new} =
v_{\neg ck}^{old} - \eta_{m}\frac{\partial E}{\partial v_{\neg ck}^{old}} =
v_{\neg ck}^{old}-\eta_{m}(1-\a_{c}+\a_{\neg c})
\frac{\a_{\neg c}}{\s_{\neg ck}^{old^{2}}}(x_{k}-v_{\neg ck}^{old})$\hf\lb
\hs*{2.5 cm}(C) $\s_{ck}^{new} =
\s_{ck}^{old} - \eta_{s}\frac{\partial E}{\partial \s_{ck}^{old}} =
\s_{ck}^{old}+\eta_{s}(1-\a_{c}+\a_{\neg c})
\frac{\a_{c}}{\s_{ck}^{old^{3}}}(x_{k}-v_{ck}^{old})^{2}$\hf\lb
\hs*{2.5 cm}(D) $\s_{\neg ck}^{new} =
\s_{\neg ck}^{old} - \eta_{s}\frac{\partial E}{\partial \s_{\neg ck}^{old}}=
\s_{\neg ck}^{old}-\eta_{s}(1-\a_{c}+\a_{\neg c})
\frac{\a_{\neg c}}{\s_{\neg ck}^{old^{3}}}(x_{k}-v_{\neg ck}^{old})^{2}$\hf\lb
\hs*{2 cm}End For \hf\lb
\hs*{1 cm}End For \hf\lb
\hs*{1 cm}Compute the  error   $E_{t}$ for the new rule base $R^{t}$.\hf\lb
\hs*{1 cm}Compute the misclassification $M_{t}$ for $R^{t}$.\hf\lb
\hs*{1 cm}If $M_{t} > M_{t-1}$ or $E_{t} > E_{t-1}$ \hf\lb
\hs*{2 cm}then\hf\lb
\hs*{2 cm}$\eta_{m} \leftarrow (1-\e)\eta_{m}$\hf\lb
\hs*{2 cm}$\eta_{s} \leftarrow (1-\e)\eta_{s}$\hf\lb
\hs*{2 cm}$R^{t} \leftarrow R^{t-1}$\hf\lb
\hs*{3 cm}/*{\em If the error is increased, then possibly the learning coefficients are \hf\lb
\hs*{3 cm}too large. So, decrease the learning coefficients, and retain $R_{t-1}$.}*/\hf\lb
\hs*{1 cm}If $M_{t} = 0$ or $E_{t} = 0$ \hf\lb
\hs*{2 cm}then Stop.\hf\lb
\hs*{1 cm}$t\leftarrow t+1$\hf\lb
End While\hf\lb

{\em End}

At the end of the rulebase tuning   we get  the final rulebase $R_{final}$
which is expected to give a very low error rate.

Since a Gaussian membership function is extended to infinity, for
any data point all rules will be fired to some extent. In our
implementation, if the firing strength is less than a threshold,
$\epsilon$ $(\approx 0.01)$, then the rule is not assumed to be
fired. Thus, under this situation, the rulebase extracted by the
system may not be complete with respect to the training data.
 This can also happen when we use membership functions with
triangular or trapezoidal shapes. This is not a limitation
but a distinct advantage,
although for the data sets we used, we did not encounter such a situation.
 If  no rule is fired by a  data
point, then that point can be thought of as an outlier.
If this happens for some test data, then that will  indicate
an observation not close enough to the training data and consequently
no conclusion should be made about such test points.

Though product is a valid T-norm and has some attractive mathematical properties,
its use is conceptually somewhat unattractive. To illustrate the point let us
consider a rule having two atomic clauses in its antecedent. If the
two clauses have truth values $a$ and $b$, then intuitively the
antecedent is satisfied at least to the extent of
$\min(a,b)$. However, if product is used as the conjunction operator, we
always have $ab \leq \min(a,b)$. Thus we always under-determine the importance
of the rule. This does not cause any problem for non-classifier fuzzy systems
as defuzzification operator usually performs some kind of  normalization
with respect to the firing strength. But in classifier type applications a decision
may appear to be taken with very low confidence, when actually it is  not the case.
For example, if each antecedent clause is satisfied to the extent 0.9 and
there are 10 antecedent clauses, the firing strength becomes $0.9^{10} = 0.3487$!
Thus to avoid the use of the product and at the same time to be
able to apply calculus to derive update rules  we use a soft-min operator.

The {\bf soft-match} of $n$ positive number
$x_1,x_2,...,x_n$ is defined by
$$SM(x_1,x_2,...,x_n,q) = \left\{\frac{(x_1^q+x_2^q+...+x_n^q)}{n}\right\}^{1/q}.$$
where $q$ is any real number.
$SM$ is known as an aggregation operator with upper bound of value 1 when
$x_i \in [0,1] \forall i$. This operator is
used by different authors \cite{dyck,chinmoy} for different purposes. It is
easy to see that
$$\lim_{q\to \infty}SM(x_1,x_2,...,x_n,q) = \max(x_1,x_2,...,x_n)$$
and
$$\lim_{q\to -\infty}SM(x_1,x_2,...,x_n,q) = \min(x_1,x_2,...,x_n).$$
Thus we define the softmin operator as the soft match operator with a sufficiently
negative value of the parameter $q$. The firing strength of the r-th rule
computed using softmin is
$$\alpha_r({\bf x}) = \left\{\frac{\sum_{j=1}^{j=p}(\mu_{rj}(x_j; v_{rj}, \sigma_{rj}))^q}{p}\right\}^{1/q}.$$
In the current study we use $q = -10.0$.

Using the same error function as in the previous section we derive the rule update equations
L, M, N, and O bellow.The tuning algorithm remains the same except equations A, B, C and D are
replaced by L, M, N, and O respectively.

(L) $v_{ck}^{new} =
v_{ck}^{old} - \eta_{m}\frac{\partial E}{\partial v_{ck}^{old}} =
v_{ck}^{old}+\eta_{m}(1-\a_{c}+\a_{\neg c})
\frac{\a_{c}}{\sum_{j=1}^{j=p}\mu_{cj}^q}\frac{\mu_{cj}^q}{\s_{ck}^{old^{2}}}(x_{k}-v_{ck}^{old})$\hf\lb
(M) $v_{\neg ck}^{new} =
v_{\neg ck}^{old} - \eta_{m}\frac{\partial E}{\partial v_{\neg ck}^{old}} =
v_{\neg ck}^{old}-\eta_{m}(1-\a_{c}+\a_{\neg c})
\frac{\a_{\neg c}}{\sum_{j=1}^{j=p}\mu_{\neg cj}^q}\frac{\mu_{\neg cj}^q}{\s_{\neg ck}^{old^{2}}}(x_{k}-v_{\neg ck}^{old})$\hf\lb
(N) $\s_{ck}^{new} =
\s_{ck}^{old} - \eta_{s}\frac{\partial E}{\partial \s_{ck}^{old}} =
\s_{ck}^{old}+\eta_{s}(1-\a_{c}+\a_{\neg c})
\frac{\a_{c}}{\sum_{j=1}^{j=p}\mu_{cj}^q}\frac{\mu_{cj}^q}{\s_{ck}^{old^{3}}}(x_{k}-v_{ck}^{old})^{2}$\hf\lb
(O) $\s_{\neg ck}^{new} =
\s_{\neg ck}^{old} - \eta_{s}\frac{\partial E}{\partial \s_{\neg ck}^{old}}=
\s_{\neg ck}^{old}-\eta_{s}(1-\a_{c}+\a_{\neg c})
\frac{\a_{\neg c}}{\sum_{j=1}^{j=p}\mu_{\neg cj}^q}\frac{\mu_{\neg cj}^q}{\s_{\neg ck}^{old^{3}}}(x_{k}-v_{\neg ck}^{old})^{2}$\hf\lb

The use of softmin is also consistent with our perception of AND
connective in the antecedent parts of the fuzzy rules. Since for
a reasonably big negative value (such as -10.0) of $q$ the softmin
makes a very good approximation of {\it min}, when the trained system is used for
testing/deployment, we can directly use {\it min} to reduce the computational
overhead.

\section{Context-Sensitive Inferencing}
The use of softmin operator opens up a host of theoretical possibilities.
The important fact to be noted is that the softmin operator is just one member of
the family of aggregation operators generated by the {\bf soft-match} operator
for $q\in [-\infty,\infty]$. The family of operators covers a large spectrum
from $minimum$ to $maximum$ including the $average$ (for $q=1$). Figure 1 shows
that softmin varies from the minimum of its arguments to their maximum via the
average.

In fuzzy rule based systems for pattern classification tasks, we use
rules of the form
\bc
$R_i:$ If $x_1$ is $A_{i1}$ AND $\cdots$ AND $x_p$ is $A_{ip}$ then class is $j$,
\ec

Even if we use a tunable conjunction operator typically
 all rules in a system use the same conjunction operator.
We can raise a fundamental question at this point. Is it really necessary
to have the same conjunction operator for all rules in a system? It is very difficult
to have a definite answer. A rule is considered to be a tool for reasoning in a
small area of input feature space, each rule can be
thought as a different context of reasoning. Thus drawing an analogy with the
reasoning of human experts, we can recognize the possibility that
within the same system there could be rules using different conjunction operators.
Thus a system may contain some rules for which minimum is the appropriate
conjunction operator while there could be others whose firing strength is larger
than the minimum of the membership values of atomic propositions. There could even
be some rules whose conjunction operator are closer in spirit to the maximum. This
leads to a concept called {\em context sensitive inferencing}. Human being often do
context sensitive  inferencing. Depending on the cost involved with a decision
an expert may adopt different level of conservatism in inferencing \cite{chinmoy}.

Thus while designing a scheme for rule extraction from the data, if the conjunction
operator for each rule can also be learnt from the data, the resulting system
is expected to achieve better performance.

In our present scheme we use the
softmin as the conjunction operator. As mentioned above, it can act as different
conjunction operators for different value of its parameter $q$. So we can calculate
the firing strength for the $r^{th}$ rule $R_r$ as
$$\alpha_r({\bf x}) = \left\{\frac{\sum_{j=1}^{j=p}(\mu_{rj}(x_j; v_{rj}, \sigma_{rj}))^{q_r}}{p}\right\}^{1/q_r}.$$
i.e., $q_r$ is the parameter for conjunction operator corresponding to the $r^{th}$
rule. Hence, the error function can be written as
\beq E(q_1,q_2,\cdots q_{\mid R \mid}) = \sum_{{\bf x} \in X}(1-\a_{c}+\a_{\neg c})^{2}, \eeq
where $R$ is the set of rules. Now using gradient descent we can  formulate an update scheme
for $q_i$s in addition to $v_{ij}$ and $\sigma_{ij}$s for reducing the error function.
The algorithm for tuning
consequent operators is given bellow.

{\em The conjunction operator refinement algorithm:}\hf\lb

{\em Begin}\hf\lb

Choose learning parameter $\eta_{q}$.\hf\lb
Choose a parameter reduction factor $ 0 < \e < 1 $.\hf\lb
Choose the maximum number of iteration $maxiter$.\hf\lb
Compute   the error   $E_{0}$ for the initial rule base $R^{0}$.  \hf\lb

$t \leftarrow 1$\hf\lb
While ($t \leq maxiter$) do\hf\lb
\hs*{1 cm}For each  vector ${\bf x} \in X$\hf\lb
\hs*{2 cm}Find the rules ${R}_{c}$ and ${ R}_{\neg c}$
 using  $\a_{c}$ and $\a_{\neg c}$.\hf\lb
\hs*{2 cm}Modify the parameters $q_c$ and $q_{\neg c}$ of rules
${R}_{c}$ and ${R}_{\neg c}$ respectively as follows .\hf\lb

\hs*{2 cm}$q_{c}^{new} = q_{c}^{old} - \eta_{q}\frac{\partial
E}{\partial q_{c}^{old}} = q_{c}^{old}+\eta_{q}(1-\a_{c}+\a_{\neg
c}) \frac{\a_{c}}{q_c}
\left(\frac{\sum_{j=1}^{j=p}\mu_{cj}^{q_c}\ln
\mu_{cj}}{\sum_{j=1}^{j=p}\mu_{cj}^{q_c}}-\ln \a_c\right)$\hf\lb

\hs*{2 cm}$q_{\neg c}^{new} = q_{\neg c}^{old} -
\eta_{q}\frac{\partial E}{\partial q_{\neg c}^{old}} = q_{\neg
c}^{old}+\eta_{q}(1-\a_{c}+\a_{\neg c}) \frac{\a_{\neg c}}{q_{\neg
c}} \left(\frac{\sum_{j=1}^{j=p}\mu_{\neg cj}^{q_{\neg c}}\ln
\mu_{\neg cj}}{\sum_{j=1}^{j=p}\mu_{\neg cj}^{q_{\neg c}}}-\ln
\a_{\neg c} \right)$\hf\lb

\hs*{1 cm}End For \hf\lb

\hs*{1 cm}Compute the  error   $E_{t}$ for the modified rule base $R^{t}$.\hf\lb
\hs*{1 cm}If $E_{t} > E_{t-1}$ \hf\lb
\hs*{2 cm}then\hf\lb
\hs*{2 cm}$\eta_{q} \leftarrow (1-\e)\eta_{q}$\hf\lb
\hs*{2 cm}$R^{t} \leftarrow R^{t-1}$\hf\lb
\hs*{3 cm}/*{\em If the error is increased, then possibly the learning coefficients are \hf\lb
\hs*{3 cm}too large. So, decrease the learning coefficients, and retain $R_{t-1}$.}*/\hf\lb
\hs*{1 cm}$t \leftarrow t+1$\hf\lb
End While\hf\lb

{\em End}

The initial set of rules used in this algorithm is the set obtained from the tuning
algorithm described in the previous section. Thus, the earlier tuning scheme finds a
suitable context for each rule using softmin and then we tune the inferencing scheme
depending on the context. In the new algorithm, unlike the previous two,
the stress is put on the
reduction of total error as defined in eq. (8). The algorithm starts with the same softmin
operator for all rules and then the operator for each rule is tuned separately. The change
in performance in terms of misclassification may not be dramatic, because as our
experiments show that in most of the cases the conjunction operators become more
or less equal to minimum. However, it was seen in the experiments that for a particular data
irrespective of the
initial value of the $q_i$s, the final sets of $q_i$s obtained are very much similar.
This indicates the possibility of existence of a natural set of conjunction operator
for a particular dataset.

\section{Implementation and results}

Sat-image1 is prepared from a four channel Landsat image consisting of 6435 pixels. So
there are 6435 vectors in ${\cal R}^4$. There are six (c=6) types of
land-cover as shown in Table 1. This is a benchmark data set and available
on the web at \cite{satim}. Performance of many classifiers for this
data set can be found in \cite{ludmila}.

The  Sat-image2 is of size $512 \times 512$ pixels captured by
seven detectors operating in different spectral bands from
Landsat-TM3. Each of the detectors generates an  image with pixel
values varying from 0 to 255. The $512 \times 512$
 ground truth data
provide the actual distribution of classes of objects captured in the image. From this
data we produce the labeled data set with each pixel represented by a 7-dimensional
feature vector and a class label. Each dimension of a feature vector comes
from one channel and the class label comes from the ground
truth data. The class distribution of the samples is given in Table 2.

 Each data set  $X$  is partitioned  into $X_{Tr}$ and $X_{Te}$ such that
$X=X_{Tr} \cup X_{Te}$ and $X_{Tr} \cap X_{Te} = \phi$. Some
benchmark results are available for both Sat-image1 and Sat-image2.
 For Sat-image1 the
associated   training-test
partition is also available \cite{satim}. We use this partition as our first
partition and
generated three more random partitions keeping the same number of
 representations
from different classes in the training and test sets.
In case of  Sat-image2 the benchmark results are
generated using 200 random samples from each class to constitute the
training data. We used the same philosophy to generate four such random
partitions.

\begin{table}
\caption{Different classes and their frequencies for Sat-image1}
\bc
\begin{tabular}{|l|r|}\hline
Land-cover types & Frequencies \\ \hline
 Red soil & 1533\\
Cotton crop & 703\\
Gray soil & 1358\\
Damp gray soil & 626\\
Soil with  & 707\\
vegetation stubble &\\
Very deep & 1508\\
gray soil &\\ \hline
Total &6435\\ \hline
\end{tabular}
\ec
\end{table}

For Sat-image1,
$X_{Tr}=500, X_{Te}=5935$. The first partition used is the same
one as used in \cite{ludmila}. The number of pixels of different
land-cover types
in the training set are :   Red soil = 104; Cotton crop = 68;
 Gray soil = 108;  Damp gray soil = 47; Soil with vegetation
 stubble =  58;
 Very deep gray soil = 115. The other partitions are randomly
generated  keeping  the same number  of pixels of different
land-cover types in the training and test sets as in the
first partition.

\begin{table}
\caption{Classes and their frequencies in the Sat-image2.}
\begin{center}
\begin{tabular}{|l|r|}  \hline
Classes         &Frequencies \\ \hline
Forest          &176987 \\
Water           &23070 \\
Agriculture     &26986 \\
Bare ground     &740 \\
Grass           &12518 \\
Urban area      &11636 \\
Shadow          &3197 \\
Clouds          &358 \\ \hline
Total           &262144 \\ \hline
\end{tabular}
\end{center}
\end{table}

We divide this section in two parts. In the first part ({\it context-free inferencing})
 we describe  performance
of two types of fuzzy rule based classifiers. The first type uses
{\bf product} as the conjunction operator. The other type uses {\bf softmin}
as the conjunction operator with a constant $q$ value for all rules in a classifier.
In the current study all these classifiers use $q = -10$.
In the second part ({\it context sensitive inferencing}) we report results using
the context sensitive inferencing scheme,
i.e., different conjunction operator for different rules.

\subsection{Performance of the classifiers with Context-Free Inferencing}

\begin{table}
\caption{Performance of the fuzzy rule based  classifiers designed with context-free
reasoning scheme for 4 different partitions of Sat-image1
 data set.}
\begin{center}
\begin{tabular}{|l|l|l|l|l|r|}  \hline
Partition   &Number of   &\multicolumn{4}{c|}{\% of Error}  \\ \cline{3-6}
Number      &Rules      &\multicolumn{2}{c|}{Product rules} &\multicolumn{2}{c|}{Softmin rules}\\ \cline{3-6}
        &       &Trng.  &Test       &Trng.  &Test   \\ \hline
1       &27     &12.8\% &15.51\%    &10.8\% &15.58\% \\ \hline
2       &26     &12.2\% &15.6\%     &10.0\% &16.16\% \\ \hline
3       &25     &16.0\% &15.26\%    &12.8\% &15.92\%\\ \hline
4       &19     &12.0\% &17.1\%     &9.4\%  &16.29\%\\ \hline

\end{tabular}
\end{center}
\end{table}

\begin{table}
\caption{Performances of fuzzy rule based  classifiers designed with context-free
reasoning scheme for different training sets for Sat-image2}
\begin{center}
\begin{tabular}{|l|c|c|c|c|c|r|} \hline
Training    &No. of     &$k_{w}$    &\multicolumn{2}{c|}{Product rules} &\multicolumn{2}{c|}{Softmin rules}\\ \cline{4-7}
Set     &rules      &       &Error Rate in  &Error Rate in      &Error Rate in  &Error Rate in \\
        &       &       &Training Data  &Whole Image        &Training Data  &Whole Image \\ \hline
1.      &29     &5.0        &19.3\%     &13.8\%         &12.0\%     &13.6\% \\ \hline
2.      &24     &6.0        &14.4\%     &13.6\%         &14.3\%     &14.47\% \\ \hline
3.      &24     &5.0        &16.5\%     &13.7\%         &12.0\%     &13.03\% \\ \hline
4.      &25     &4.0        &16.0\%     &13.8\%         &12.6\%     &12.5\% \\ \hline
\end{tabular}
 \end{center}
\end{table}

\begin{table}

\caption{The rules for classes 1, 2 and 3 with corresponding fuzzy
sets for Sat-image 2. These rules are obtained using the training
set 1. }
\begin{center}
\begin{tabular}{|l|l|l|} \hline
Class  &Rule No.   &Fuzzy sets in form of $(\mu_i,\s_i)$ tuples, $i=1,2,\cdots ,7$ \\
\hline
1      &2           &(65.7,11.9), (23.6,6.8), (22.8,10.2), (50.9,19.8)\\
        &           &(41.7,16.2), (130.3,6.1), (13.9,12.4) \\\cline{2-3}
        &6          &(67.5,11.4), (24.8,7.0), (25.4,10.6), (56.5,22.1) \\
        &           &(56.5,19.7), (131.9,4.8), (18.3,4.2) \\\cline{2-3}
        &8          &(68.8,11.7), (27.4,8.7), (30.7,12.2), (57.0,19.4)\\
        &           &(62.1,24.8), (126.5,8.5), (7.9,11.9)\\\hline
2       &4          &(65.5,12.1), (22.0,7.5), (17.1,9.4),(9.1,2.2)\\
        &           &(4.9,6.7), (128.7,6.9), (2.1,8.2)\\ \hline
3       &7          &(67.1,13.3), (24.9,7.7), (20.6,11.3), (45.4,21.7)\\
        &           &(59.1,17.6), (130.7,5.7), (27.1,4.3)\\ \cline{2-3}
        &9          &(66.2,8.5), (25.0,9.3), (24.7,10.1), (53.9,20.2)\\
        &           &(77.3,12.3), (137.1,6.4), (27.1,14.5)\\ \cline{2-3}
        &14         &(71.5,13.3), (27.9,7.7), (23.8,9.9), (88.4,17.3)\\
        &           &(89.5,16.9), (137.2,12.1), (41.8,13.7)\\ \cline{2-3}
        &15         &(71.9,22.3), (26.3,22.7), (27.2,45.0), (72.7,39.5)\\
        &           &(84.7,46.6), (133.8,12.0), (29.7,40.2)\\ \hline

\end{tabular}
 \end{center}
\end{table}

Table 3 depicts the performance of the fuzzy rule based classifier
on Sat-image1.
Both fuzzy rule based classifiers show consistently almost similar performances.
In all cases the softmin based classifier shows some improvement in training
error. While for test sets the product based classifier shows slightly better
performance in three cases and the softmin based classifier is better in the
remaining case.
 Sat-image1 has been
extensively studied in \cite{ludmila}. Comparing our results with
the results in \cite{ludmila} we find that our classifier
outperforms Multi-layer
Perceptron (MLP) network and produces comparable results as that by
Radial Basis Function (RBF) network. For example, in \cite{ludmila}
the test error reported using a MLP is 23.08\% while that
by RBF networks varied between 14.52\% - 15.52\%

In Table 4 we include the results of four different random partitions of
Sat-image2. It shows an excellent performance of the rule based classifiers.
Both classifiers
show consistent and comparable performances. For all partitions the softmin
based classifiers show lower training errors while for the whole data
the product based classifier performs better in only one case.  Figure 2
shows one channel of Sat-image2 while Fig. 3 depicts the ground truth
values where each class is represented by a distinct gray level. Fig. 4
shows a typical classification result (corresponding to the first
 partition) which
is almost identical to Fig. 3.

The same data set (Sat-Image 2) has been used by Kumar  et al. \cite{kumar2}
  in a comparative study using several classification techniques. The
best result obtained by them using a {\em fuzzy integral} based scheme gives a
classification rate 78.15\%. {\em In our case, even the worst performance
is about 5\% better than the results in} \cite{kumar2}.

In Table 5 we show the rules obtained
for  classes 1, 2 and 3 of Sat-image 2 using the training set 1.
Column 1 of Table 5 shows the class number while
column 2 lists the rule identification number.
Column 3 describes the rules using
the membership functions in the antecedent part of the
rules. Since it is a seven dimensional data, each rule
involves seven atomic antecedent clauses (fuzzy sets).
Each fuzzy set is represented by a 2-tuple $(\mu_i,\s_i)$, where
$\mu_i$ and $\s_i$ are the center and spread of a Gaussian membership function.
Thus for rule number 2, the tuple (65.7,11.9) represents a clause
"gray value from channel 1 is CLOSE to 65.7" where the fuzzy set
CLOSE to 65.7 is represented by a Gaussian function with center at
65.7 and spread 11.9. Inspection of the parameters of
rules for class 1 reveals that between rule 2 and rule 6 all features
change, but features 4, 5 and 7 changes significantly.
Comparing rules 6 and 8 we find that features 3, 5, 6 and 7 change
significantly. This indicates that each rule represents distinct areas
in the feature space.

Table 5 also suggests that data from class 2 probably form a nice cluster
in the feature space that can be modeled by just a single
rule. This is further confirmed by the fact that
the spread of the membership functions for features 2,4,5,6 and 7 are
relatively small.

Similarly, for class 3, different rules model different areas
in the feature space. For rules 14 and 15 although $\mu_i$'s for
features 1 and 2 do not change much, the $\mu_i$'s for other features
change considerably between the two rules. Depending on the complexity
of the class structure, the required number of rules  also changes.
The number of rules for a given class may also vary depending on
the training set used. However, this variation across the training
sets is not much. For example, the number of rules for class 1
obtained using 4 training sets are 3, 2, 3 and 4 respectively.
Similar results are obtained for other classes too. This
indicates good robustness of the proposed rule generation and
tuning algorithm.

\subsection{Performance of the Classifiers with context sensitive inferencing}
\begin{table}
\caption{Performance of the rule based  classifiers designed with context-sensitive
reasoning scheme for 4 different partitions of Sat-image1
 data set.}
\begin{center}
\begin{tabular}{|l|l|l|r|}  \hline
Partition   &Number of   &\multicolumn{2}{c|}{\% of Error}  \\ \cline{3-4}
Number      &Rules  &Trng.  &Test   \\ \hline
1       &27     &11.0\% &15.58\% \\ \hline
2       &26     &9.8\%  &16.19\%  \\ \hline
3       &25     &12.8\% &15.96\% \\ \hline
4       &19     &9.6\% &16.22\%  \\ \hline

\end{tabular}
\end{center}
\end{table}

\begin{table}
\caption{Classification performances designed with context-sensitive
reasoning scheme for 4 different training sets for Sat-image2}
\begin{center}
\begin{tabular}{|l|c|c|r|} \hline
Training    &No. of     &Error Rate in  &Error Rate in \\
Set     &rules      &Training Data  &Whole Image \\ \hline
1.      &29     &11.87\%    &13.5\% \\ \hline
2.      &24     &14.6\%     &14.45\% \\ \hline
3.      &24     &12.19\%    &13.18\% \\ \hline
4.      &25     &12.75\%    &13.1\% \\ \hline
\end{tabular}
 \end{center}
\end{table}

To study the effect of context sensitive inferencing scheme on the
classifiers we performed several experiments. First we tuned the
set of rules obtained from the context tuning (context free
inferencing) stage using softmin operators with fixed value of
$q(=-10.0)$  parameters. In this experiment the values of $q_i$s
for all the rules are initially set to -10.0, while the learning
parameter $\eta_q=500$ and $maxiter = 200$.

The performances of the context sensitive classifiers after conjunction operator
tuning are summarized
in Tables 6 and 7 for Sat-image1 and Sat-image 2 respectively. It can be observed
that the performances in terms of classification rate do not improve significantly.
However, some improvement in terms of training error as defined by (8) is
observed in all cases. It is further seen that after tuning, the $q_i$s for all
rules remain negative, which makes every rule to use approximately the
minimum as the conjunction operator. This is hardly surprising, since the initial rules
are already context tuned with fixed value of $q=-10.0$ for all rules. So it
indicates that the context tuning with {\it fixed} $q$ has developed the rule set
to correspond to an energy minima.

To investigate this issue further, we tuned the same initial set of rules (i.e.,
the context tuned rules with fixed $q=-10.0$) with different initial values of
$q_i$s, namely, 1.0 and 5.0. All classifiers, with the same training data
set has shown strong tendencies of convergence to similar set of conjunction operators,
irrespective of the initial value of the $q_i$s. We present the result
of the classifiers designed with the partition 1 of satimage-1 and satimage-2 with
initial values of $q_i$s 1.0 and 5.0 in Table 8. The results for initial $q_i$ -10.0 are
also included in this table for comparison.
Column 2 of Table 8 (and 9) shows the initial value of q for all rules in the rule base
when the q-tuning for every rule (to realize {\it context sensitive inferencing}) starts.

\begin{table}
{\small \caption{Performance analysis of context sensitive
inferencing for rule based classifiers (when the initial rules
were context tuned with q=-10.0 for all rules)}
\begin{center}
\begin{tabular}{|l|c|c|c|c|c|c|c|} \hline
Training    &Initial    &\multicolumn{2}{c|}{Training)} &\multicolumn{2}{c|}{Training}  &\multicolumn{2}{c|}{Misclassification in test} \\
data        &$q_i$      &\multicolumn{2}{c|}{Error ($E$)}&\multicolumn{2}{c|}{Misclassification}&\multicolumn{2}{c|}{data or whole image} \\ \cline{3-8}
        &       &Initial  &Final        &Initial    &Final      &Initial    &Final\\ \hline
Sat-image1  &-10.0      &176.26   &175.9        &54 (10.8\%)    &55 (11.0\%)    &925 (15.6\%)   &917 (15.4\%)\\ \cline{2-7}
        &1.0        &359.71   &176.02       &72 (14.4\%)    &55 (11.0\%)    &1000 (16.8\%)  &915 (15.4\%)\\ \cline{2-7}
        &5.0        &468.09   &176.28       &116 (23.2\%)   &54 (10.8\%)    &1536 (25.9\%)  &908 (15.3\%)\\ \hline
Sat-image2  &-10.0      &688.07   &685.06       &192 (12.0\%)   &189 (11.8\%)   &35657 (13.6\%) &35202 (13.4\%)\\ \cline{2-7}
        &1.0        &1283.42  &724.98       &274 (17.1\%)   &202 (12.6\%)   &37241 (14.2\%) &35585 (13.5\%)\\ \cline{2-7}
        &5.0        &1487.01  &684.86       &414 (25.8\%)   &190 (11.8\%)   &50330 (19.2\%) &35082 (13.4\%)\\ \hline
\end{tabular}
 \end{center}
}
\end{table}

Table 8 shows that for both cases with initial $q_i$s 1.0 and 5.0,
though the initial value of error and misclassification rates are
quite high, after the tuning they are much lower and very close to
those corresponding to classifiers with initial $q_i$ -10.0 for
the respective data set. This fact is reflected for the training
sets as well as the test sets. It was further observed that for
all the initial setting of $q_i$ the final values of $q_i$ become
negative making the operator more or less equivalent to the
$minimum$ operator. They are observed to attain values close in
magnitude also. A small but interesting exception occurs in case
of Satimage-2 (training set 1). Figure 5 shows the bar diagram of
the tuned $q$ values of the rules for Satimage-2. In the figure
each group of three bars shows the tuned $q$ values for a
particular rule for three different initial values of $q = -10.0$
(white), 1.0 (gray) and 5.0 (black) respectively. For rule 1 with
initial $q_1 = 1.0$, it can be observed that $q_1$ remains 1.0. We
analyzed the case and found that the rule was not fired at all and
the value remained the same as the initial value. The results in
tables 8 and 9 suggest that the context tuning using softmin has
successfully reduced the error and developed a set of rules such
that the optimal performance of the classifiers can be obtained
with minimum-like operators. Thus, it may be noted that for these
two data sets direct use of minimum does an equally good job.


To investigate the validity of the above we carried out another set of experiments. Here
we used the initial rule set as obtained from the prototype generation stage, i.e.,
the rules are {\it no more context-tuned}. We tuned these rules for initial $q_i$s -10.0, 1.0
and 5.0 respectively. The results of tuning for the partition 1 of Satimage1 and Satimage2
are summarized in Table 9. As evident from the results, in all the cases both initial
and final performances are worse than those in the previous experiment. However, the results
also  reveal that the conjunction operator (context sensitive) tuning have produced
substantial improvement of performances in all cases. Thus it shows that if the initial rule
base is not very good, the context sensitive tuning can result in substantial improvement in
performance.The bar plots of the tuned $q$
values of the rules for satimage-1 and satimage-2 are shown in figures 6 and 7
respectively. It can easily be discerned from the figures that the final set of
$q_i$ values, unlike the previous experiment, differ substantially for different
initial values of the $q_i$s. This clearly indicates that the system in this
case attains different error minima in $q_i$-space for different initialization.

\begin{table}
{\small
\caption{Performance analysis of context sensitive inferencing for rule based classifiers
(when the initial rules were not context tuned)}
\begin{center}
\begin{tabular}{|l|c|c|c|c|c|c|c|} \hline
Training    &Initial    &\multicolumn{2}{c|}{Training)} &\multicolumn{2}{c|}{Training}  &\multicolumn{2}{c|}{Misclassification in test} \\
data        &$q_i$      &\multicolumn{2}{c|}{Error ($E$)}&\multicolumn{2}{c|}{Misclassification}&\multicolumn{2}{c|}{data or whole image} \\ \cline{3-8}
        &       &Initial  &Final        &Initial    &Final      &Initial    &Final\\ \hline
Sat-image1  &-10.0      &258.42   &235.39       &70 (14.0\%)    &67 (13.4\%)    &1035(17.4\%)   &958(16.14\%)\\ \cline{2-7}
        &1.0        &345.41   &245.15       &74 (14.8\%)    &67 (13.4\%)    &1097 (18.5\%)  &941 (15.8\%)\\ \cline{2-7}
        &5.0        &471.22   &246.02       &116 (23.2\%)   &68 (13.6\%)    &1680 (28.3\%)  &1042 (17.5\%)\\ \hline
Sat-image2  &-10.0      &1198.47   &1081.71     &426 (26.6\%)   &398 (24.8\%)   &68070 (25.9\%) &62873 (23.9\%)\\ \cline{2-7}
        &1.0        &1280.22  &1096.01      &417 (26.0\%)   &382 (23.8\%)   &63645 (24.2\%) &56464 (21.5\%)\\ \cline{2-7}
        &5.0        &1446.36  &1143.70      &472 (29.5\%)   &404 (25.2\%)   &74299 (28.3\%) &57174 (21.8\%)\\ \hline
\end{tabular}
 \end{center}
}
\end{table}

\section{Conclusion}
Fuzzy rule based classifiers are capable of dealing with
data having different variances for different features and also different variances for
different classes. In such  a system a rule represents a region in the input
space, which we call the {\bf context} of the rule. Thus a fuzzy rule based classifier
is inherently capable of detecting outliers also. Here we proposed
a scheme for designing a fuzzy rule based classifier. It starts with generation
of a set of prototypes. Then these prototypes are converted into fuzzy rules.
The rules are then tuned with respect to their context. We have developed two
variants of the context tuning algorithm, one for rules using product as conjunction
operator and the other is used when the conjunction operator is softmin.

The rule based classifier is then used to classify multi-spectral
satellite images and the performance obtained is excellent.
The classifier performance may further be improved using features other than
gray labels. Also use of other T-norms can alter the performance of the
classifier. We are currently investigating all these possibilities.

If softmin is used as the conjunction operator, then using different values of
the $q$ parameters of softmin for different rules, the rules can be
made to use effectively different conjunction operators. Such a scheme
is called {\em context sensitive inferencing}. We have developed an algorithm
for tuning the $q$ parameters on per-rule basis.

Our experimental result suggests that if the context tuning (with fixed $q$ value
for all rules) is carried out properly then subsequent context sensitive tuning
offers marginal (if any) improvement. However, if the rules are not context
tuned, the context sensitive inferencing can improve the performance of the
classifier substantially.

Some other facts also came to light from these experiments.
The results of $q$ tuning of the context-tuned rules suggest possible
existence of a deep error minima with respect to the $q$ parameters to which
the system reached for different initializations. But the results for the
rule sets without context-tuning indicates that the system lands up in different
minima for different initializations.

In light of above we can suggest two possible utility for using context sensitive
reasoning for classification: (1) if the context tuning is not good enough, the
performance may be enhanced with context sensitive reasoning and (2) the results
of tuning the $q_i$s
with different initializations may be used for judging the quality of a
rulebase as a whole. This aspect is currently under further investigation.

\pagebreak

\bfg
\centerline{\psfig{figure=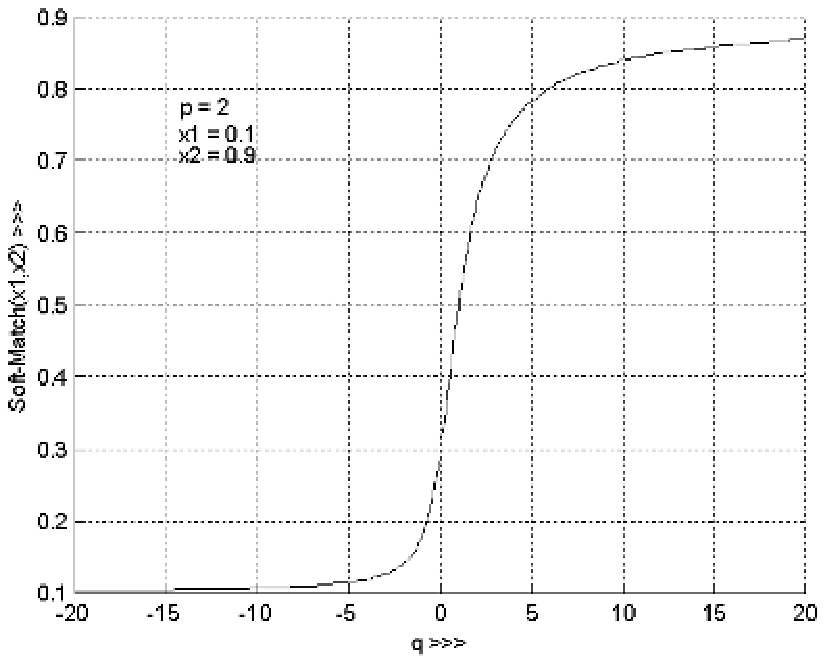,height=4cm}}
\caption{Plot of soft-match operator against $q$.}
\efg

\bfg
\centerline{\psfig{figure=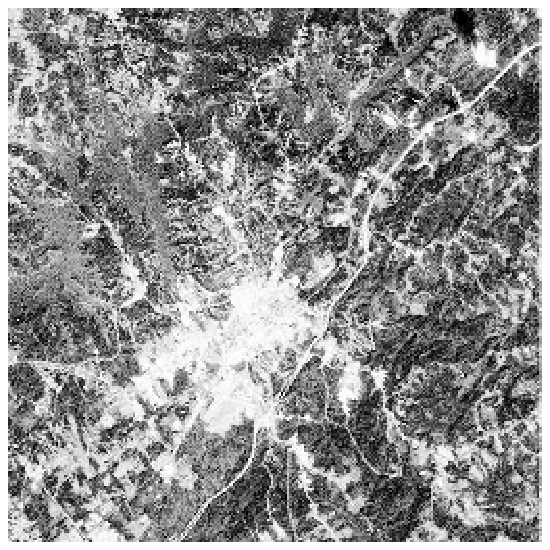,height=5cm }}
\caption{Band-1 of Sat-image2 (After histogram equalization).}
\efg

\bfg \centerline{\psfig{figure=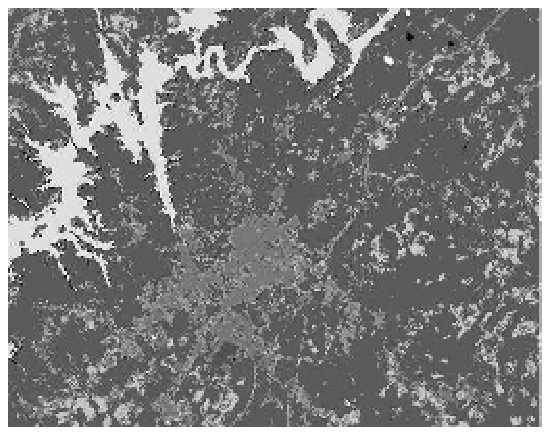,height=5cm }} \caption{
The ground truth for Sat-image2. The classes are represented by
different colors.} \efg \pagebreak

\bfg \centerline{\psfig{figure=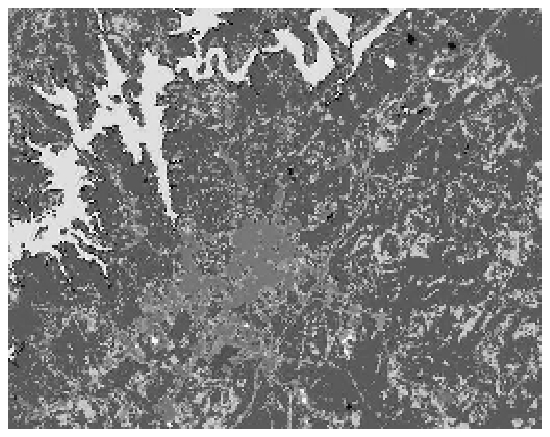,height=5cm }} \caption{
The classified image for Sat-image2. The classes are represented
by different colors.} \efg

\bfg
\centerline{\psfig{figure=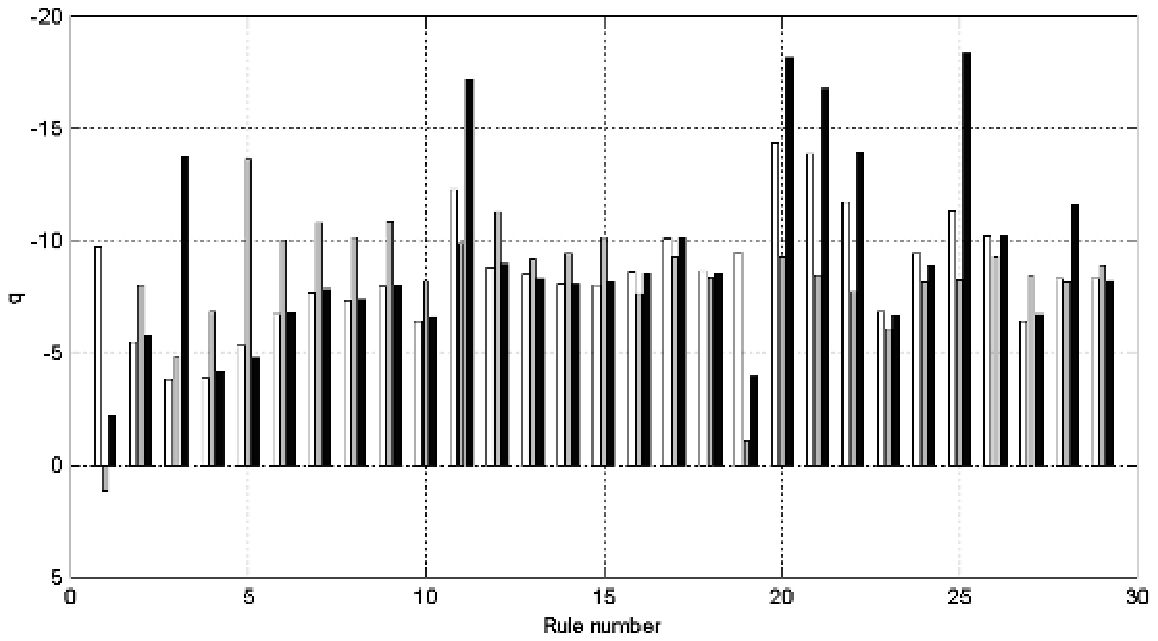,height=5cm}}
\caption{Bar diagram of $q_i$s of the rules for Sat-image2 for initial $q_i$ -10, 1 and 10
(when the initial rules were context tuned with q=-10.0 for all rules).}
\efg

\bfg
\centerline{\psfig{figure=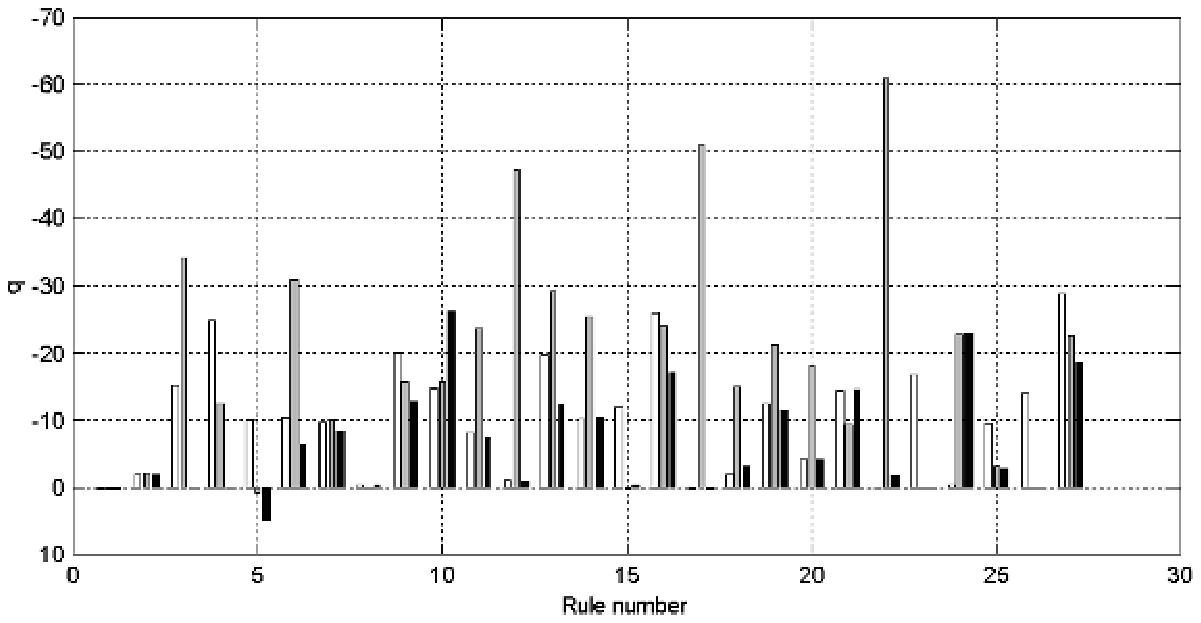,height=5cm}}
\caption{Bar diagram of $q_i$s of the rules for Sat-image1 for initial $q_i$ -10, 1 and
5 (when the initial rules were not context tuned).}
\efg

\bfg
\centerline{\psfig{figure=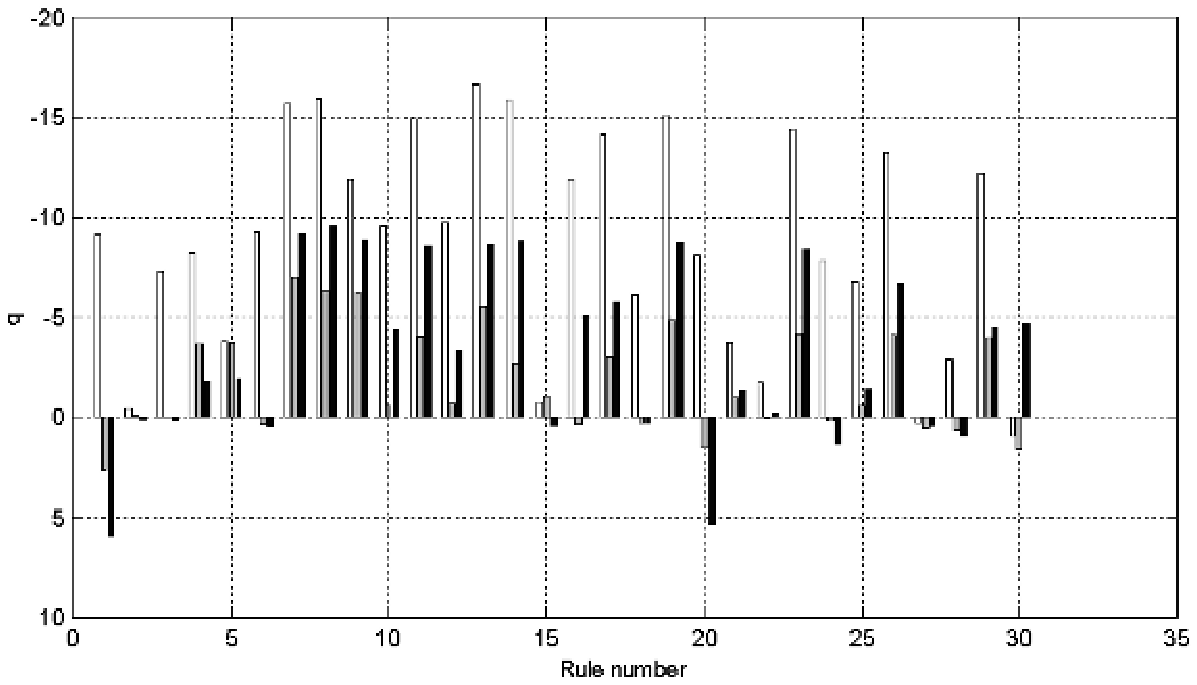,height=5cm}}
\caption{Bar diagram of $q_i$s of the rules for Sat-image2 for initial $q_i$ -10, 1 and
5 (when the initial rules were not context tuned).}
\efg

\end{document}